\title{DeepKSPD: Learning Kernel-matrix-based SPD Representation for Fine-grained Image Recognition}
\author{
        Melih Engin, Lei Wang\thanks{Corresponding author (leiw@uow.edu.au)}, Luping Zhou \\
                School of Computing and Information Technology\\
        University of Wollongong\\
        Wollongong, NSW 2500, \underline{Australia}
            \and
        Xinwang Liu\\
        School of Computer\\
        National University of Defense Technology\\
        Changsha, Hunan 410073, \underline{China}
}
\date{}

\documentclass[12pt]{article}
\usepackage{amsmath,amssymb,bm}
\usepackage{graphicx}
\usepackage[table,x11names]{xcolor}
\newcommand\mywidth{25mm}

\begin{document}
\maketitle

\begin{abstract}
Being symmetric positive-definite (SPD), covariance matrix has traditionally been used to represent a set of local descriptors in visual recognition. Recent study shows that kernel matrix can give considerably better representation by modelling the nonlinearity in the local descriptor set. Nevertheless, neither the descriptors nor the kernel matrix is deeply learned. Worse, they are considered separately, hindering the pursuit of an optimal SPD representation. This work proposes a deep network that jointly learns local descriptors, kernel-matrix-based SPD representation, and the classifier via an end-to-end training process. We derive the derivatives for the mapping from a local descriptor set to the SPD representation to carry out backpropagation. Also, we exploit the {Dalecki\v{i}-Kre\v{i}n formula} in operator theory to give a concise and unified result on differentiating SPD matrix functions, including the matrix logarithm to handle the Riemannian geometry of kernel matrix. Experiments not only show the superiority of kernel-matrix-based SPD representation with deep local descriptors, but also verify the advantage of the proposed deep network in pursuing better SPD representations for fine-grained image recognition tasks.
\end{abstract}

\section{Introduction}
To deal with image variations, modern visual recognition usually models the appearance of an image by a set of local descriptors. They evolve from early filter bank responses, through traditional local invariant features, to the activation feature maps of recent deep convolutional neural networks (CNNs). During the course, how to represent a set of local descriptors to obtain a global image representation has always been a central issue. Among the best known methods in this line of research are the bag-of-features (BoF) model~\cite{DBLP:conf/iccv/SivicZ03}, sparse coding~\cite{DBLP:conf/cvpr/WangYYLHG10}, vector of locally aggregated descriptors (VLAD)~\cite{DBLP:conf/cvpr/JegouDSP10}, and Fisher vector encoding~\cite{DBLP:journals/ijcv/SanchezPMV13}. In recent years, representing a set of descriptors with a covariance matrix has attracted increasing attention. It characterizes the pairwise correlation of descriptor components presented in a set, and is generally called SPD representation since covariance matrix is symmetric positive-definite. This representation is robust to noisy descriptors and independent of the cardinality of a descriptor set. Also, it does not need a large number of images to generate common bases for encoding, and can therefore be individually applied to single images. In the past few years, covariance-matrix-based SPD representation has been employed in a variety of visual recognition tasks including the recognition of texture, face and object~\cite{DBLP:conf/cvpr/JayasumanaHSLH13}, the classification of image set~\cite{DBLP:conf/cvpr/WangGDD12}, and so on.

A recent progress on SPD representation is to model the nonlinear information in a set of descriptors. As reported in~\cite{DBLP:conf/iccv/WangZZTL15}, directly using a kernel matrix to represent a descriptor set demonstrates its superiority. Given a set of $d$-dimensional descriptors, a $d\times{d}$ kernel matrix is computed with a predefined kernel function, where each entry is the kernel value between the realization of two descriptor components in this set. This method effectively models the nonlinear correlation among these descriptor components. The kernel function can be flexibly chosen to extract various nonlinear relationship, and the covariance is just a special case using a linear kernel. The resulting kernel-matrix-based SPD representation maintains the same size as its covariance-matrix-based counterpart, but produces considerable improvement on recognition performance.

Nevertheless, this kernel-matrix-based SPD representation~\cite{DBLP:conf/iccv/WangZZTL15} is only developed upon traditional local descriptors like the pixel intensities or the Gabor filter responses of a textured or facial image. Its potential with deep local descriptors on image recognition has not been explored in the literature and therefore remains unclear. Another more critical issue, which is the main focus of this work, is that the local descriptors and the kernel matrix in the existing SPD representation are detached. In other words, they cannot effectively negotiate with each other to obtain an optimal representation for the ultimate goal of classification. To address these two issues, this work completely builds the kernel-matrix-based SPD representation built upon deep local descriptors and benchmarks it against the state-of-the-art image recognition methods. More importantly, we develop a deep network called DeepKSPD to jointly learn the deep local descriptors, the kernel-matrix-based SPD representation, and the classifier. This is achieved by an end-to-end training process between input images and class labels. The presence of kernel matrix computation in the proposed deep network complicates the backpropagation process. Also, to make the resulting SPD representation better work with the classifier, a matrix logarithm function is usually required to map the kernel matrix from Riemannian geometry to Euclidean geometry. In this work, we derive all the matrix derivatives involved in the mapping from a local descriptor set to the kernel-matrix-based SPD representation to fulfill the backpropagation algorithm for the proposed deep network. Also, by exploiting the {Dalecki\v{i}-Kre\v{i}n formula} in operator theory~\cite{Daleckii-Krein,Bhatia:2015:PDM:2838858}, we provide a concise and unified result on the derivative of the functions on SPD matrices, in which matrix logarithm is a special case. Together, these produce a backpropagation algorithm that could deal with a deep network with various kernel-matrix-based SPD representations. 

Experimental study is conducted on multiple benchmark datasets, especially on fine-grained image recognition, to demonstrate the efficacy of the proposed DeepKSPD framework. First, in contrast to the existing kernel-matrix-based representation built upon traditional local descriptors, we demonstrate the superiority of the kernel-matrix-based SPD representation built upon deep local descriptors. On top of that, we further demonstrate the advantage of the proposed end-to-end trained DeepKSPD network in jointly learning the local descriptors and the kernel-matrix-based SPD representation. As will be shown, our DeepKSPD network achieves the overall highest classification accuracy on these benchmark datasets, when compared with the related deep learning based methods. 

\section{Related Work}\label{sec:related-work}

Let ${\bm X}_{d\times{n}}=[{\bm x}_{1},{\bm x}_{2},\cdots,{\bm x}_{n}]$ denote a data matrix, in which each column contains a local descriptor ${\bm x}_{i}~({\bm x}_{i}\in{\mathcal R}^{d})$, extracted from an image. In the days when local invariant features such as SIFT are popularly used, methods like BoF, VLAD, and Fisher vector have been developed to encode and pool these descriptors to obtain a global image representation. VLAD and Fisher vector methods have recently been applied to the deep local descriptors collected from the activation feature maps of deep CNNs, demonstrating promising image recognition performance~\cite{DBLP:conf/eccv/GongWGL14,DBLP:conf/cvpr/CimpoiMV15}. These methods usually need a sufficient number of images to train a set of common bases (e.g., cluster centers or Gaussian mixture models) for the encoding step.   

The SPD representation takes a different approach. It traditionally computes a $d\times{d}$ covariance matrix over ${\bm X}$ as ${\bm\varSigma}=\bar{\bm X}{\bar{\bm X}}^{T}$ (or simply ${\bm X}{\bm X}^{T}$), where $\bar{\bm X}$ denotes the centered ${\bm X}$. Originally, this covariance matrix is proposed as a region descriptor, for example, characterizing the covariance of the color intensities of pixels in a local image patch. In the past several years, it has been employed as a promising global image representation in a number of visual recognition tasks. Recent research in this line aims to model the nonlinear information in a set of descriptors. The approach proposed in~\cite{DBLP:conf/cvpr/HarandiSP14} implicitly maps each descriptor ${\bm x}_{i}~(i=1,2,\cdots,n)$ onto a kernel-induced feature space and computes a covariance matrix therein. Nevertheless, this results in a high (or even infinite) dimensional covariance matrix that is difficult to manipulate explicitly or computationally. The other approach~\cite{DBLP:conf/iccv/WangZZTL15} proposes to directly compute a kernel matrix ${\bm K}$ over ${\bm X}$ as follows. Let ${\bm f}_{j}$ denote the $j$th row of ${\bm X}$, consisting of the $n$ realizations of the $j$th component of ${\bm x}$. The $(i,j)$th entry of ${\bm K}$ is calculated as $k({\bm f_i},{\bm f_j})$, with a predefined kernel function $k$ such as a Gaussian kernel. In this way, the nonlinear relationship among the $d$ components can be effectively and flexibly extracted. The resulting kernel matrix ${\bm K}$ maintains the size of $d\times{d}$ and is more robust against the singularity issue caused by small sample. It is easy to see that covariance matrix is a special case in which $k$ reduces to a linear kernel function. As reported in~\cite{DBLP:conf/iccv/WangZZTL15}, this kernel-matrix-based SPD representation achieves considerably better performance than its covariance counterpart and that proposed in~\cite{DBLP:conf/cvpr/HarandiSP14} on multiple visual recognition tasks. 

Both covariance and kernel matrices are SPD and have a Riemannian geometry. In order to work with the common classifiers that assume a Euclidean geometry, a variety of operations have been developed in the literature. Among them, the matrix logarithm operation, $\log(\cdot)$, may be the most commonly used one due to its simplicity and efficacy~\cite{MRM:MRM20965}. Conceptually, it can be viewed as mapping an SPD matrix from the underlying manifold to its tangent space in which Euclidean geometry can be applied. In practice, after the matrix ${\bm K}$ is obtained, the matrix $\log({\bm K})$ will be computed and then reshaped into a long vector to be fed into a classifier. 

Research on integrating the SPD representation \textit{with} deep local descriptors or even \textit{into} deep networks is still in its very early stage but has demonstrated both theoretical and practical values. In the recent work of Bilinear CNN~\cite{DBLP:conf/iccv/LinRM15,lin2017bilinear}, an outer product layer is applied to combine the activation features maps from two CNNs, and this produces clear improvement in fine-grained visual recognition. This outer product essentially leads to a covariance matrix (in the form of ${\bm X}{\bm X}^{T}$) when the two CNNs are set as the same. Another work in~\cite{DBLP:conf/iccv/IonescuVS15} trains a deep network for image semantic segmentation, in which the covariance-matrix-based SPD representation is used to represent a set of local descriptors. 

Nevertheless, to the best of our survey, all the existing few works on SPD representation in deep learning focus on the covariance-matrix-based SPD representation. None of them has considered the kernel-matrix-based one, which can produce significantly better recognition performance. To address this issue, we develop a deep network focusing on the kernel-matrix-based SPD representation and jointly learn this representation with deep local descriptors. Also, the work in~\cite{DBLP:conf/iccv/IonescuVS15} derives the derivations of the matrix logarithm function from the scratch. Although instructive, it does not connect this derivation with the operator theory on positive definite matrix~\cite{Bhatia:2015:PDM:2838858}. In this work, by establishing this interesting link, we not only readily obtain the derivatives for matrix logarithm (and other general SPD matrix functions) in a much concise way, but can also gain more insights by accessing the vast knowledge in that field for future research. 

At the end, it is worth noting that in this work the kernel matrix is integrated into deep neural networks as a representation of a set of local descriptors collected from the activation feature maps. This is fundamentally different from the recent works that develop new CNNs with reproducing kernels, supervised convolutional kernel networks, and deep kernel learning models~\cite{MairalKHS14,Mairal16,WilsonHSX16b}. 
\begin{figure*}[t]
\centering
\includegraphics[width=0.9\textwidth]{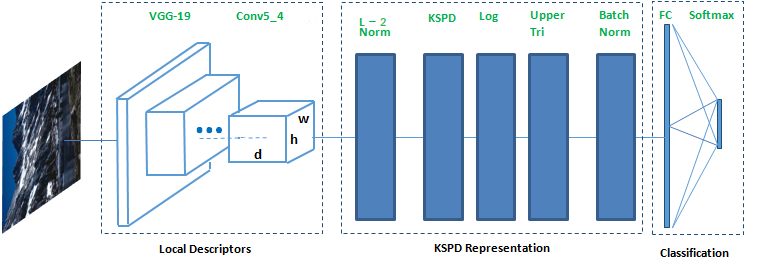}
\caption{The structure of the proposed DeepKSPD network}\label{fig:structure} 
\end{figure*}
\section{The proposed network DeepKSPD}\label{sec:DeepKSPD}
The proposed network DeepKSPD consists of three blocks, as shown in Fig.~\ref{fig:structure}. The leftmost block maps an input image into a set of deep local descriptors. Since we deal with visual recognition, any convolutional neural network can be used. It generates a set of activation feature maps for an image, from which a set of deep local descriptors are collected. In this work we employ the commonly used VGG-$19$ network pre-trained on the ImageNet dataset. The rightmost block includes the commonly used fully connected and softmax layers to produce the posteriori probability for each class. In between is the KSPD block that contains the layers related to the kernel-matrix-based representation and the matrix logarithm. In specific, the input of the KSPD block is the output of the last convolutional layer (conv$5$\_$4$) of the VGG-$19$ network. In this way, the input consists of $d$ activation feature maps of the size of $w\times{h}$. These feature maps are reshaped along the depth dimension $d$, and this gives rise to the matrix ${\bm X}_{d\times{n}}$ with $n=w\times{h}$. Afterwards, the kernel matrix ${\bm K}_{d\times{d}}$ is computed with ${\bm X}$. It pools the $n$ deep local descriptors by capturing the pairwise nonlinear relationship among the $d$ feature maps. Following that is the matrix logarithm layer to handle the Riemannian geometry of SPD matrix and this produces the matrix ${\bm H}=\log({\bm K})$. Since ${\bm H}$
is symmetric, a layer that extracts the upper triangular and diagonal entries of ${\bm H}$ is deployed next to avoid redundancy. We observe that normalized KSPD representations usually perform better. Therefore, an $L_2$ normalization and a batch normalization layer are added at the two ends of the KSPD block, respectively. 

\section{End-to-end training of DeepKSPD}

\subsection{Derivatives between $\bf{\textit X}$ and the kernel matrix $\bf{\textit K}$}

Recall that ${\bm X}_{d\times{n}}$ denotes a set of local descriptors. Considering that Gaussian kernel is commonly used in the literature and that it is used in~\cite{DBLP:conf/iccv/WangZZTL15} to demonstrate the advantage of the kernel-matrix-based representation, we exemplify the proposed DeepKSPD with a Gaussian kernel and focus on this case to derive the derivatives. Other kernels such as polynomial kernel can be dealt with in a similar way. 

\begin{figure}[h]
	\includegraphics[width=1.0\textwidth]{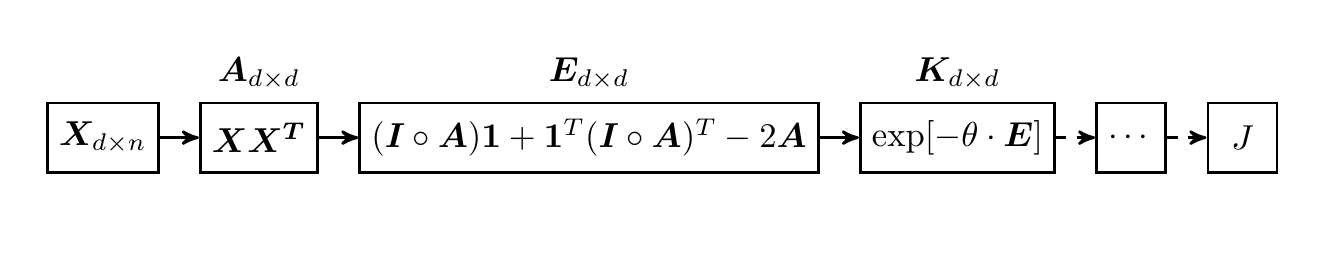}
	\caption{Illustration of the mapping from ${\bm X}$ to ${\bm K}$ (a Gaussian kernel function is used).}\label{fig:flow} 
\end{figure}

Let ${\bm I}_{d\times{d}}$ and ${\bm 1}_{d\times{d}}$ denote an identity matrix and a matrix of $1$s. Let $\circ$ denote the entrywise product (Hadamard product) of two matrices, and $\exp[\cdot]$ denote an exponential function applied to a matrix in an entrywise manner. In this way, the Gaussian kernel matrix ${\bm K}$ computed on ${\bm X}$ can be expressed as
{\small
\begin{equation}
 {\bm K} = \exp\left[-\theta\cdot\left(({\bm I}\circ{{\bm X}{\bm X}^T}){\bm 1}+{\bm 1}^T({\bm I}\circ{{\bm X}{\bm X}^T})^T-2{\bm X}{\bm X}^T\right)\right],
\end{equation}
}where $\theta$ is the width of the Gaussian kernel. This expression is illustrated in Fig.~\ref{fig:flow}. 

Let $J$ denote the objective function to be optimized by the DeepKSPD network. By temporarily assuming that the derivative $\frac{\partial{J}}{{\partial{\bm K}}}$ has been known (will be resolved in the next section), we now work out the derivative $\frac{\partial{J}}{{\partial{\bm X}}}$ and $\frac{\partial{J}}{{\partial\theta}}$. Note that $J$ is a composition of functions applied to ${\bm X}$ and it can be equally expressed as a function of each of the intermediate variables as follows.  
{\small
\begin{equation}\label{eqn:compos}
J({\bm X})=J_{1}({\bm A})=J_{2}({\bm E})=J_{3}({\bm K}),
\end{equation}
}where ${\bm A}$, ${\bm E}$, and ${\bm K}$ are defined as
{\small
\begin{align}
\begin{aligned}
&{\bm A}={\bm X}{\bm X}^T,\quad
\\
&{\bm E}=\left(({\bm I}\circ{\bm A}){\bm 1}+{\bm 1}^T({\bm I}\circ{\bm A})^T-2{\bm A}\right),\quad
\\
&{\bm K}=\exp[-\theta\cdot{\bm E}]
.
\end{aligned}
\end{align}
}%
Following the rules for differentiation, the following relationship can be obtained
{\small
\begin{align}\label{eqn:diff}
\begin{aligned}
&\delta{\bm A}=\left(\delta{\bm X}\right){\bm X}^T+{\bm X}\left(\delta{\bm X}\right)^T,~ 
\\
&\delta{\bm E}=({\bm I}\circ\delta{\bm A}){\bm 1}+{\bm 1}^T({\bm I}\circ\delta{\bm A})^T-2\delta{\bm A},~
\\
&\delta{\bm K}=\left({-\theta}{\bm K}\right)\circ{\delta{\bm E}}.
\end{aligned}
\end{align}
}%
Furthermore, it is known from the differentiation of a scalar-valued matrix function that
{\small
\begin{equation}\label{eqn:scala-diff}
 \delta{J}=\left\langle\mathrm{vec}\left(\frac{\partial{J_3}}{\partial{\bm K}}\right),\mathrm{vec}({\delta{\bm K}})\right\rangle = \mathrm{trace}\left(\left(\frac{\partial{J_3}}{\partial{\bm K}}\right)^T{\delta{\bm K}}\right),
\end{equation}
}%
where $\mathrm{vec}(\cdot)$ denotes the vectorization of a matrix and $\langle\cdot,\cdot\rangle$ denotes the inner product. Combining this result with $\delta{\bm K}=\left({-\theta}{\bm K}\right)\circ{\delta{\bm E}}$ in Eq.(\ref{eqn:diff}) and using the identity that $\mathrm{trace}({\bm A}^{T}({\bm B}\circ{\bm C}))=\mathrm{trace}(({\bm B}\circ{\bm A})^T{\bm C})$, we can obtain
{\small
\begin{equation}\label{eqn:dK-dE}
\begin{split}
 \delta{J}=\mathrm{trace}\left(\left(\frac{\partial{J_3}}{\partial{\bm K}}\right)^T{\delta{\bm K}}\right) & =\mathrm{trace}\left(\left(-\theta{\bm K}\circ\frac{\partial{J_3}}{\partial{\bm K}}\right)^T{\delta{\bm E}}\right)
  \\ & =\mathrm{trace}\left(\left(\frac{\partial{J_2}}{\partial{\bm E}}\right)^T{\delta{\bm E}}\right).
 \end{split}
\end{equation}
}%
The last equality holds because from Eq.(\ref{eqn:compos}) we know that $\delta{J}$ can also be written as $\mathrm{trace}\left(\left(\frac{\partial{J_2}}{\partial{\bm E}}\right)^T{\delta{\bm E}}\right)$. Noting that Eq.(\ref{eqn:dK-dE}) is true for any $\delta{\bm E}$, we can therefore derive that
{\small
\begin{equation}\label{eqn:J2-J3}
\frac{\partial{J_{2}}}{{\partial{\bm E}}} = \left({-\theta}{\bm K}\right)\circ\frac{\partial{J_{3}}}{{\partial{\bm K}}}.
\end{equation}
}%
Repeating the above process by using the relationship of $\delta{\bm E}$ and $\delta{\bm A}$ and that of $\delta{\bm A}$ and $\delta{\bm X}$ in Eq.(\ref{eqn:diff}), we can further have (proof is provided in Appendix)
{\small
\begin{align}
\begin{aligned}
& \frac{\partial{J_{1}}}{{\partial{\bm A}}} = {\bm I}\circ\left(\left(\frac{\partial{J_{2}}}{{\partial{\bm E}}}+\left(\frac{\partial{J_{2}}}{{\partial{\bm E}}}\right)^T\right){\bm 1}^T\right)-2\frac{\partial{J_{2}}}{{\partial{\bm E}}};
\\
\quad
& \frac{\partial{J}}{{\partial{\bm X}}} = \left(\frac{\partial{J_{1}}}{\partial{\bm A}}+\left(\frac{\partial{J_{1}}}{\partial{\bm A}}\right)^T\right){\bm X}.
\end{aligned}
\end{align}
}%
In addition, the derivative $\frac{\partial{J}}{{\partial{\theta}}}$ can be obtained as
{\small
\begin{equation}
\frac{\partial{J}}{{\partial{\theta}}} = \mathrm{trace}\left(\left(\frac{\partial{J_3}}{{\partial{\bm K}}}\right)^T\left(-{\bm K}\circ{\bm E}\right)\right).
\end{equation}
}%
Therefore, when $\frac{\partial{J_3}}{{\partial{\bm K}}}$ is available, we can work out $\frac{\partial{J}}{{\partial{\bm X}}}$ and $\frac{\partial{J}}{\partial{\theta}}$ according to the above results. 

\subsection{Derivatives of the matrix logarithm on the kernel matrix $\bf{\textit K}$}

Now, to obtain $ \frac{\partial J_3}{\partial \bm K} $ we deal with the matrix logarithm operation between ${\bm K}$ and $J$, which can be written as
{\small
\begin{equation}\label{eqn:log-process}
J({\bm X})=J_{4}({\bm H})=J_{4}(\log({\bm K})).
\end{equation}
}%
Note that $\frac{\partial{J_4}}{{\partial{\bm H}}}$ is ready to obtain because it only involves the classification layers like fully connected layer, softmax regression and cross-entropy computation. The key issue is to obtain $\frac{\partial{\bm H}}{{\partial{\bm K}}}$. In the following we introduce the \textit{Dalecki\v{i}-Kre\v{i}n formula}~\cite{Daleckii-Krein} to give a concise and unified result on differentiating SPD matrix functions, of which the matrix logarithm is a special case. 

\noindent{\textbf{Theorem 1} (pp.$60$, \cite{Bhatia:2015:PDM:2838858})} \textit{Let ${\mathbb M}_{d}$ be the set of $d\times{d}$ real symmetric matrices. Let $I$ be an open interval and ${\mathbb M}_{d}(I)$ is the set of all real symmetric matrices whose eigenvalues belong to $I$. Let $C^{1}(I)$ be the space of continuously differentiable real functions on $I$. Every function $f$ in $C^{1}(I)$ induces a differentiable map from ${\bm A}$ in ${\mathbb M}_{d}(I)$ to $f({\bm A})$ in ${\mathbb M}_{d}$. Let $Df_{\bm A}(\cdot)$ denote the derivative of $f({\bm A})$ at ${\bm A}$. It is a linear map from ${\mathbb M}_{d}$ to itself. When applied to ${\bm B}\in{\mathbb M}_{d}$, $Df_{\bm A}(\cdot)$ is given by the {Dalecki\v{i}-Kre\v{i}n formula} as
{\small
\begin{equation}\label{eqn:Df_A_B}
 Df_{\bm A}({\bm B}) = {\bm U}\left({\bm G}\circ\left({\bm U}^T{\bm B}{\bm U}\right)\right){\bm U}^T,
\end{equation}
}%
where ${\bm A}={\bm U}{\bm D}{\bm U}^T$ is the eigendecomposition of ${\bm A}$ with ${\bm D}=\mathrm{diag}(\lambda_1,\cdots,\lambda_{d})$, and $\circ$ is the entrywise product. The entry of the matrix ${\bm G}$ is defined as
{\small
\begin{equation}\label{eqn:g_ij}
g_{ij} = \left\{\begin{array}{l} 
\frac{f(\lambda_i)-f(\lambda_j)}{\lambda_i-\lambda_j}~\mbox{if}~\lambda_i\neq{\lambda_j}\\
f'(\lambda_i),~\mbox{otherwise.}
\end{array} \right. 
\end{equation}
}%
}This theorem indicates that for a matrix function $f(\cdot)$ applied to ${\bm A}$, perturbing ${\bm A}$ by a small amount ${\bm B}$ will vary $f({\bm A})$ by the quantity $Df_{\bm A}({\bm B})$ in Eq.(\ref{eqn:Df_A_B}), where the variation is in the sense of the first-order approximation. Now we show how to derive the functional relationship between $\frac{\partial{J_4}}{{\partial{\bm H}}}$ and $\frac{\partial{J_3}}{{\partial{\bm K}}}$ based on Theorem 1. According to Eq.(\ref{eqn:compos}) and following the argument in Eq.(\ref{eqn:scala-diff}), we have 
{\small
\begin{equation}\label{eqn:dH-dK}
 \delta{J}=\mathrm{trace}\left(\left(\frac{\partial{J_4}}{\partial{\bm H}}\right)^T{\delta{\bm H}}\right)=\mathrm{trace}\left(\left(\frac{\partial{J_3}}{\partial{\bm K}}\right)^T{\delta{\bm K}}\right).
\end{equation}
}%

Applying the {Dalecki\v{i}-Kre\v{i}n formula}, we can explicitly represent $\delta{\bm H}$ to be a function of $\delta{\bm K}$ as
{\small
\begin{equation}
{\delta{\bm H}} = Df_{\bm K}(\delta{\bm K}) = {\bm U}\left({\bm G}\circ\left({\bm U}^T{\delta{\bm K}}{\bm U}\right)\right){\bm U}^T. 
\end{equation}
}%
Replacing $\delta{\bm H}$ in Eq.(\ref{eqn:dH-dK}) with the above result and again applying the properties of $\mathrm{trace}({\bm A}^T{\bm B})$, the relationship between $\frac{\partial{J_4}}{{\partial{\bm H}}}$ and $\frac{\partial{J_3}}{{\partial{\bm K}}}$ can be derived in a similar way as in Eqs.(\ref{eqn:dK-dE}) and (\ref{eqn:J2-J3})
{\small
\begin{equation}\label{eqn:pk-ph}
 \frac{\partial{J_3}}{\partial{\bm K}} = {\bm U}\left({\bm G} \circ \left({\bm U}^T{\frac{\partial{J_4}}{\partial{\bm H}}}{\bm U}\right)\right) {\bm U}^T.
\end{equation}
}%
where ${\bm U}$ and ${\bm G}$ are obtained from the eigendecomposition of ${\bm K}={\bm U}{\bm D}{\bm U}^T$. The matrix logarithm $f({\bm K})\triangleq\log({\bm K})$ is now just a special case in which $g_{ij}$ in Eq.(\ref{eqn:g_ij}) is computed as $\frac{\log\lambda_i-\log\lambda_j}{\lambda_i-\lambda_j}$ when $i\neq{j}$ and $\lambda_i^{-1}$ otherwise. 

The work in~\cite{DBLP:conf/iccv/IonescuVS15} derives the derivative of the matrix logarithm from the scratch with the basic facts of matrix differentiation, which is instructive. However, as previously mentioned, that work does not connect this derivative with the well-established {Dalecki\v{i}-Kre\v{i}n formula}. To consolidate this connection and link with the work in~\cite{DBLP:conf/iccv/IonescuVS15}, we prove the following proposition.

\noindent{\textbf{Proposition 1} \textit{The functional relationship obtained in~\cite{DBLP:conf/iccv/IonescuVS15} shown in Eq.(16) (with the notation in this work for consistency) is equivalent to that in Eq.(\ref{eqn:pk-ph}) obtained by this work.  
{\small
\begin{equation}\label{eqn:pk-ph-apdx}
\begin{aligned}
 \frac{\partial{J_3}}{\partial{\bm K}} & = {\bm U}\left\{\left(\tilde{\bm G} \circ \left(2{\bm U}^{T}\left({\frac{\partial{J_4}}{\partial{\bm H}}}\right)_{sym}{\bm U}\log({\bm D})\right)\right)\right. \\ & + \left. \left({\bm D}^{-1}\left({\bm U}^T{\frac{\partial{J_4}}{{\partial{\bm H}}}}{\bm U}\right)\right)_{diag}\right\}{\bm U}^T,
\end{aligned}
\end{equation}
}%
where ${\bm K}={\bm U}{\bm D}{\bm U}^T$; $\tilde{g}_{ij}=(\lambda_i-\lambda_j)^{-1}$ when $i\neq{j}$ and zero otherwise; ${\bm A}_{diag}$ means the off-diagonal entries of ${\bm A}$ are all set to zeros; and ${\bm A}_{sym}$ is defined to represent $({\bm A}+{\bm A}^T)/2$.}

\noindent{\textbf{Proof.} Note that $\frac{\partial{J_{4}}}{\partial{\bm H}}$ is symmetric because ${\bm H}$ is symmetric. Therefore, $\left(\frac{\partial{J_{4}}}{\partial{\bm H}}\right)_{sym}$ just equals $\frac{\partial{J_{4}}}{\partial{\bm H}}$. In this way, Eq.(\ref{eqn:pk-ph-apdx}) can be written as
{\small
\begin{equation}\label{eqn:J3-K}
\begin{aligned}
&\frac{\partial{J_3}}{\partial{\bm K}} = {\bm U}\left\{\left(\tilde{\bm G}^{\top}\circ\left(2{\bm U}^{\top}{  \left(\frac{\partial{J_{4}}}{\partial{\bm H}}\right){\bm U}\log({\bm D})}\right)\right)\right. \\ & + \left.\left({\bm D}^{-1}\left({\bm U}^{\top}\frac{\partial{J_{4}}}{\partial{\bm H}}{\bm U}\right)\right)_{diag}\right\}{\bm U}^{\top}\\~\nonumber
& \left[\mbox{Define}~{\bm Z}=\frac{\partial{J_{4}}}{\partial{\bm H}}~\mbox{for the sake of clarity}\right]\\~\nonumber
&={\bm U}\left\{\left(\tilde{\bm G}^{\top}\circ\left(2{\bm U}^{\top}{{\bm Z}{\bm U}\log({\bm D})}\right)\right)\right. \\ &
+\left.\left({\bm D}^{-1}\left({\bm U}^{\top}{\bm Z}{\bm U}\right)\right)_{diag}\right\}{\bm U}^{\top}\\~\nonumber
& \left[\mbox{Define}~{\bm P}_{d\times{d}}=((\log\lambda_1){\bm 1},\cdots,(\log\lambda_d){\bm 1})\right]\\~\nonumber
&={\bm U}\left\{\left(2\tilde{\bm G}^{\top}\circ{\bm P}\right)\circ\left({\bm U}^{\top}{\bm Z}{\bm U}\right)
+{\bm D}^{-1}\circ\left({\bm U}^{\top}{\bm Z}{\bm U}\right) \right\}{\bm U}^{\top}\\~\nonumber
&={\bm U}\left\{\left(2\tilde{\bm G}^{\top}\circ{\bm P}+{\bm D}^{-1}\right)\circ\left({\bm U}^{\top}{\bm Z}{\bm U}\right) \right\}{\bm U}^{\top}\\~\nonumber
& \left[\mbox{Define}~{\bm Q}=2\tilde{\bm G}^{\top}\circ{\bm P}+{\bm D}^{-1}\right]\\~\nonumber
&={\bm U}\left\{{\bm Q}\circ\left({\bm U}^{\top}{\bm Z}{\bm U}\right) \right\}{\bm U}^{\top}.
\end{aligned}
\end{equation}
}%
Noting that $\frac{\partial{J_{3}}}{\partial{\bm K}}$ is symmetric because ${\bm K}$ is symmetric, it can be shown that
{\small
\begin{equation}\label{eqn:M+M'}
\begin{aligned}
& \frac{\partial{J_{3}}}{\partial{\bm K}} = \frac{1}{2}\left(\frac{\partial{J_{3}}}{\partial{\bm K}}+{\left(\frac{\partial{J_{3}}}{\partial{\bm K}}\right)}^{\top}\right)
={\bm U}\left\{\frac{\left({\bm Q}+{\bm Q}^{\top}\right)}{2}\circ\left({\bm U}^{\top}{\bm Z}{\bm U}\right) \right\}{\bm U}^{\top}.
\end{aligned}
\end{equation}
}%
Now let us examine the matrix of $\frac{1}{2}{\left({\bm Q}+{\bm Q}^{\top}\right)}$.
{\small
\begin{equation}
\begin{aligned}
& \frac{1}{2}{\left({\bm Q}+{\bm Q}^{\top}\right)} = (\tilde{\bm G}^{\top}\circ{\bm P})+(\tilde{\bm G}^{\top}\circ{\bm P})^{\top}+{\bm D}^{-1}\\~\nonumber
& \left[\mbox{Note that}~\tilde{\bm G}^{\top}=-\tilde{\bm G}~\mbox{according to its definition}\right]\\~\nonumber
&= (-\tilde{\bm G}\circ{\bm P})+(\tilde{\bm G}\circ{\bm P}^{\top})+{\bm D}^{-1}\\
&= \tilde{\bm G}\circ({\bm P}^{\top}-{\bm P})+{\bm D}^{-1}.
\end{aligned}
\end{equation}
}%

\noindent Noting that
{\small
\begin{equation}
\begin{aligned}
({\bm P}^{\top}-{\bm P})_{ij}
=    \begin{cases}
      {\log\lambda_i-\log\lambda_j}, & i\neq{j} \\
      0, & {i=j}
    \end{cases},
\end{aligned}
\end{equation}
}%
it can be obtained that
{\small
\begin{equation}
\begin{aligned}
\left(\tilde{\bm G}\circ({\bm P}^{\top}-{\bm P})+{\bm D}^{-1}\right)_{ij}
=    \begin{cases}
      \frac{\log\lambda_i-\log\lambda_j}{\lambda_i-\lambda_j}, & i\neq{j} \\
      {\lambda^{-1}_i}, & {i=j}
    \end{cases}=({\bm G})_{ij},
\end{aligned}
\end{equation}
}%
where ${\bm G}$ is the matrix defined in Eq.(12). Therefore it can be obtained that
{\small
\begin{equation}
\begin{aligned}
\frac{1}{2}{\left({\bm Q}+{\bm Q}^{\top}\right)}={\bm G}.
\end{aligned}
\end{equation}
}%
Combining this result with the last line of Eq.(\ref{eqn:M+M'}) in this proof gives rise to 
{\small
\begin{equation}
\begin{aligned}
 \frac{\partial{J_3}}{\partial{\bm K}} ={\bm U}\left({\bm G}~{\circ}~\left({\bm U}^{\top}\frac{\partial{J_4}}{\partial{\bm H}}{\bm U}\right)\right){\bm U}^{\top}. 
\end{aligned}
\end{equation}
}This completes the proof. \hspace{8.5cm}$\blacksquare$ 

Connecting with the results in operator theory not only facilitates the access to the derivatives of SPD matrix functions, but also provides us more insight on these functions. For example, $g_{ij}$ defined in Eq.(\ref{eqn:g_ij}) has a specific name of ``first divided difference'' of the function $f(\cdot)$, and ${\bm G}$ is called ``L\"{o}ewner matrix''~\cite{Bhatia:2015:PDM:2838858}. The positive semi-definiteness (PSD) of ${\bm G}$ guarantees the operator monotonicity of $f(\cdot)$, that is $f(\bm A)-f(\bm B)$ maintains to be PSD if ${\bm A}-{\bm B}$ is PSD. This applies to the matrix logarithm function $\log(\cdot)$ because it can be proved that ${\bm G}$ in Eq.(\ref{eqn:g_ij}) is PSD. Properties like this could be useful for the future research on SPD representations, for example, when designing a deep Siamese network that involves the difference of two SPD representations.  

\section{Experimental Result}\label{sec:exp}
There are two tasks in this experiment: i) test the performance of KSPD built upon deep local descriptors and ii) more importantly, test the performance of the proposed end-to-end learning network DeepKSPD, on the tasks of fine-grained image recognition and scene recognition, by following the literature. In the Birds dataset, bounding boxes are not used. Example images of these datasets are in Fig. 3.

\begin{figure*}[t]
\centering
\includegraphics[width=0.93\textwidth]{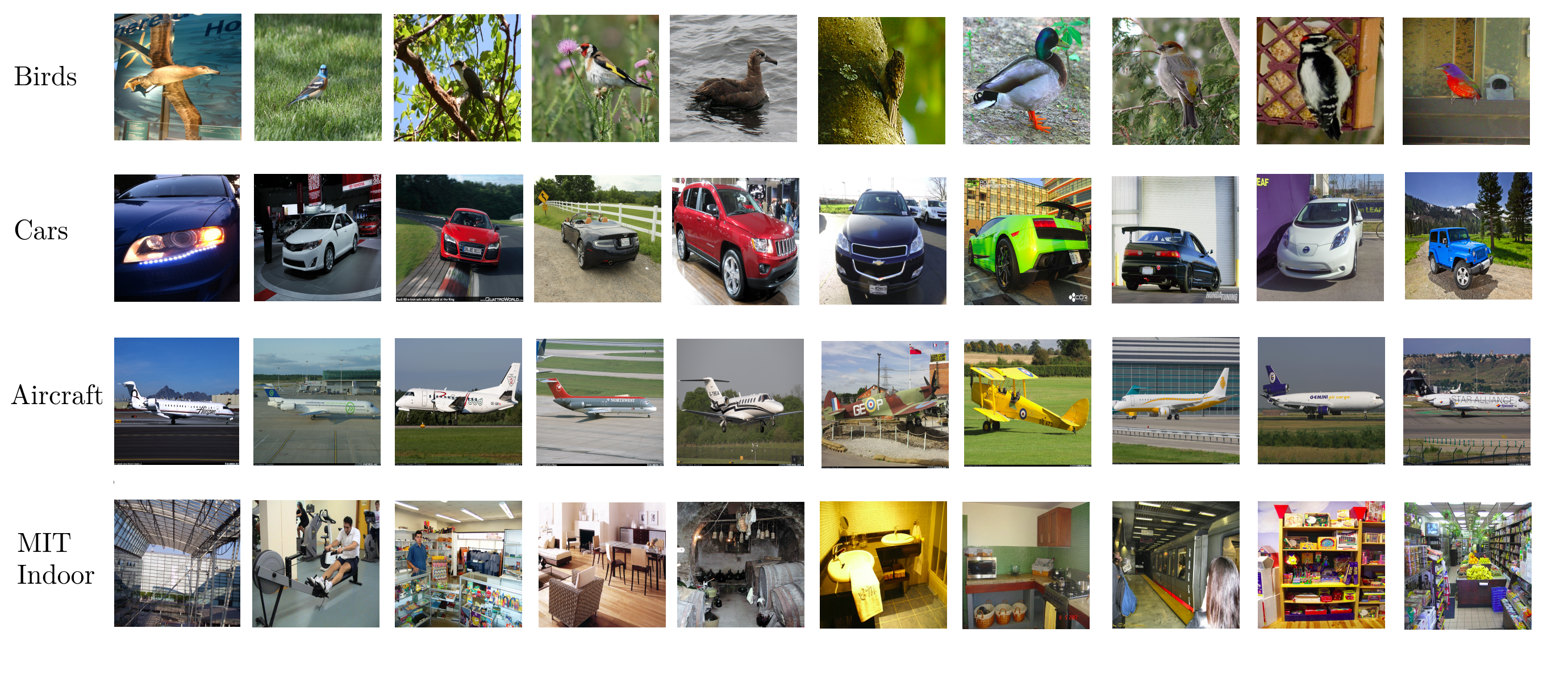}
\caption{Example images from the datasets. The top three rows correspond to the fine-grained image recognition benchmarks of Birds, Cars and Aircraft, respectively. The bottom row corresponds to the scene recognition benchmark MIT Indoor.}\label{fig:samples} 
\end{figure*}

\paragraph{Datasets}  Four benchmark data sets are employed in this experiment. For scene recognition, the MIT Indoor data set is used, which has $67$ classes with predefined $5600$ training and $1340$ test images. For fine-grained image recognition, three data sets of Cars~\cite{KrauseStarkDengFei-Fei_3DRR2013}, Birds~\cite{WelinderEtal2010}, and Aircrafts~\cite{DBLP:journals/corr/MajiRKBV13} are tested. The Cars dataset has $16185$ images from $196$ classes; the Aircrafts dataset contains $10200$ images of $100$ classes (variants). The birds dataset has $11788$ samples of $200$ bird species. All the datasets are the benchmarks widely used by the recently developed deep learning based image recognition methods. In the Birds dataset, bounding boxes are not used.

\paragraph{Setting of Proposed Methods}For the first task, we put forward a method called KSPD-VGG, which constructs kernel-matrix-based SPD representation upon the deep local descriptors extracted from VGG-$19$ pretrained on ImageNet. Specifically, the $512$ feature maps (of size $27\times27$) of the last convolutional layer of VGG-$19$ are reshaped to form $512$ vectors with the dimensions of $729$~($27\times 27$). These vectors are further used to compute the $512\times 512$ Gaussian kernel matrix ${\bm K}$. Then, after applying the matrix logarithm to the kernel matrix, only the upper triangular and diagonal parts of the resulting matrix are taken and vectorized to represent an image. The resulting KSPD representations of all images are further processed by PCA dimensionality reduction (to $4096$ dimensions), standardization (to zero mean and unit standard deviation), and  $\ell_2$ normalization. Finally, a nonlinear SVM classifier is employed to perform classification for this first task.

For the second task, the proposed DeepKSPD network is trained and tested. Note that the architecture of DeepKSPD consists of three blocks (Fig.~\ref{fig:structure}). In the local descriptor block, the network hyperparameters (e.g., the number of kernels and their sizes) are set by following VGG-$19$. In the proposed KSPD representation block, no hyperparameter needs to be preset (initial $\theta $ is set to $0.1$ for all  of the experiments). In the classification block, the size of FC layer is set as the number of classes for each data set. DeepKSPD is trained by Adaptive Moment Estimation (Adam) in mini-batch mode (with the batch-size of $20$). A two-step training procedure~\cite{BransonVBP14} is applied as good performance is observed~\cite{BransonVBP14,DBLP:conf/iccv/LinRM15}. Specifically, we first train the last layer using softmax regression for $15$ epochs, and then fine-tune the whole system. The total training epochs are $30 \sim 50$, varied with the data sets. 

\paragraph{Methods in Comparison}
We compare the proposed KSPD-VGG and DeepKSPD with a set of methods that are either comparable or competitive in the literature. They are listed in the first column in Table~\ref{Table:comparisons}, and can be roughly grouped into the following three categories. 

The first category can be deemed as feature extraction methods, to which KSPD-VGG belongs. This category also includes FV-SIFT~\cite{Perronnin:2010:IFK:1888089.1888101}, FC-VGG~\cite{DBLP:conf/iccv/IonescuVS15}, FV-VGG~\cite{DBLP:conf/cvpr/CimpoiMV15}, and COV-VGG (standing for covariance-matrix-based SPD representation). Except in FV-SIFT, the images are represented by features extracted from the pretrained deep CNN model (VGG-$19$) without fine-tuning, which allows us to better focus on the sheer effectiveness of the methods in comparison. In FC-VGG, features are extracted from the last FC layer of VGG-$19$ for classification. FV-SIFT and FV-VGG construct Fisher vectors based on local descriptors for classification. FV-SIFT uses the conventional SIFT descriptors, while FV-VGG uses the deep local descriptors from the last convolutional layer of VGG-$19$, following the literature. COV-VGG's setting is same as that of KSPD-VGG, except that a covariance matrix is constructed instead of a kernel matrix. Note that, we directly quote the results of FV-SIFT and FC-VGG from the literature, and provide our own implementation of FV-VGG, COV-VGG, and KSPD-VGG to ensure the same setting for fair comparison. 

The second category includes three end-to-end learning methods, i.e., DeepCOV, DeepKSPD (proposed) and Bilinear CNN (denoted as B-CNN)~\cite{DBLP:conf/iccv/LinRM15}. DeepCOV follows the same network architecture as the proposed DeepKSPD, but replaces the kernel matrix in the KSPD layer with a covariance matrix. DeepCOV is conceptually the same as~\cite{DBLP:conf/iccv/IonescuVS15}, but~\cite{DBLP:conf/iccv/IonescuVS15} is designed for segmentation. B-CNN is tested by using the code provided by~\cite{DBLP:conf/iccv/LinRM15}. The fine-tuned B-CNN is employed for a fair comparison with DeepCOV and DeepKSPD that involve an end-to-end training. Note that, in~\cite{DBLP:conf/iccv/LinRM15}, it shows that some engineering efforts can significantly improve the performance of B-CNN, such as augmenting the data sets by flipping images and using a separate SVM classifier instead of the softmax layer in the original deep model for classification, etc. To minimize the impacts of these engineering tricks, we switch off the image flipping component in the downloaded code, and directly perform the classification by the softmax layer as usual, same as what we do with DeepCOV and DeepKSPD. 
  
In the third category, additional methods previously reported on the involved data sets are quoted to further extend the comparison and provide a whole picture.

\paragraph{Results and Discussion}The result is summarized in Table~\ref{Table:comparisons} with the following observations.

\begin{table}[!ht]
\caption{{{Comparison of Methods {\tiny (${\dagger}$ indicates the results quoted from the literature.)}}}}\label{Table:comparisons}
\begin{minipage}[b]{1.0\linewidth}\centering
\renewcommand{\arraystretch}{1.2}
\small{
\begin{tabular}{p{5.5cm}| p{1.0cm}|p{0.7cm}| p{1.2cm}|p{1.0cm}|p{1.2cm}}
	\hline
	ACC (\%) & MIT\newline  indoor & Cars & Aircraft & Birds & Average \\\hline\hline
	
	Symbiotic Model~\cite{Chai-ICCV-13}& -- & $78.0^{\dagger}$ & $72.5^{\dagger}$ & -- & -- \\\hline
	FV-revisit~\cite{gosselin:hal-01056223}& -- & $82.7^{\dagger}$ & $80.7^{\dagger}$ & -- & -- \\\hline\hline
	FV-SIFT~\cite{Perronnin:2010:IFK:1888089.1888101} & -- & $59.2^{\dagger}$ & $61.0^{\dagger}$ & $18.8^{\dagger}$ & --\\\hline
	FC-VGG~\cite{DBLP:conf/iccv/LinRM15} & $67.6^{\dagger}$ & $36.5^{\dagger}$ & $45.0^{\dagger}$ & $61.0^{\dagger}$ & $52.5$\\\hline
	FV-VGG~\cite{DBLP:conf/cvpr/CimpoiMV15} & $73.7$ & $75.2$ & $72.7$ & $71.3^{\dagger}$ & $73.1$\\\hline
	COV-VGG &$74.2$  &$80.3$ & $81.4$ & $76$ & $77.98$ \\\hline
	\rowcolor{lightgray}  KSPD-VGG ({\bf proposed}) & $77.2$ & $83.5$ & $83.8$ & $78.5$ & $80.1$\\\hline\hline
	B-CNN~\cite{lin2017bilinear}  & $77.6$ & $87.5$ & $82.5$ & $83.5$ & $82.5$\\\hline
	DeepCOV  & $75.2$ & $86.7$ &  $81.3$ & $82.2$ & $81.35$\\\hline
	\rowcolor{lightgray}  DeepKSPD ({\bf proposed}) & $\textbf{79.6}$ & $\textbf{90.1}$ & $\textbf{86.3}$ & $\textbf{84.5}$ & $\textbf{85.13}$\\\hline\hline
\end{tabular}
}
\end{minipage}
\end{table}

First, the proposed KSPD-VGG and DeepKSPD demonstrate their effectiveness for visual recognition. On every dataset, the end-to-end learning method DeepKSPD achieves the best performance among all the methods. Overall, DeepKSPD shows superior performance over KSPD-VGG (up to $7$ percentage points on Cars) and other competitive methods, demonstrating the essentials of the end-to-end learning of kernel-matrix-based representation. 

Second, it can be seen that KSPD-based methods consistently win COV-based ones on all data sets, either based on feature extraction (KSPD-VGG vs COV-VGG) or using end-to-end learning (DeepKSPD vs DeepCOV). To ensure fair comparison, the KSPD-based and COV-based methods only differ in the SPD representation.

Third, as analyzed above, conceptually B-CNN is very close to DeepCOV when the two paths used in B-CNN are set as the same.  However, DeepCOV performs slightly worse than BCNN in the experiment (around $1\%$). Looking into this result, we find that after attaining the outer product matrix, B-CNN applies sign square-root on all entries of the matrix, rather than performing the matrix logarithm as in DeepCOV and DeepKSPD. Sign square-root can be efficiently computed by GPU, so that a much longer training procedure (up to $100$ epochs) is tolerable. However, matrix logarithm is currently implemented with CPU, whose calculation is slower than sign square-root. Therefore, we only train DeepCOV and DeepKSPD for $30 \sim 50$ epochs, and even with this setting the proposed DeepKSPD has achieved superior performance. Note that the incorporation of matrix logarithm is necessary, as it is a principled way to handle the Riemannian geometry of SPD matrix. We have observed that using more epochs and smaller learning rate, the performance of DeepKSPD and DeepCOV can be further improved, and the superiority of DeepKSPD over B-CNN will become more salient. In future, we will explore GPU-based implementation of matrix logarithm. 

Fourth, as shown, the SPD representation (being it based on an outer product, covariance, or kernel matrix) outperforms Fisher vector representation in the given visual recognition tasks. The proposed DeepKSPD also outperforms FV-VGG obtained from fine-tuned VGG-$19$. The latter attained $78.7$\% on Aircraft, on Birds $74.7$\% and $85.7$\% on Cars~\cite{DBLP:conf/iccv/LinRM15}, which is worse than $86.3$\%, $84.5$\%  and $90.1$\% achieved by DeepKSPD.

Moreover, it is worth emphasizing that this experiment focuses on comparing the core of these methods. Therefore, we minimize the engineering tricks that are detachable from the model. Certainly, e steps such as augmenting the data, fine-tuning the model for feature extraction, and applying multi-scaling, as used in the literature, can effectively improve the performance of KSPD-VGG and DeepKSPD. 

\section{Conclusion}\label{sec:conclusion}
Motivated by the recent progress on SPD representation, we develop a deep neural network that jointly learns local descriptors and kernel-matrix-based SPD representation for fine-grained image recognition. The matrix derivatives required by the backpropagation process are derived and linked to the established literature on the theory of positive definite matrix. Experimental result on benchmark datasets demonstrates the improved performance of kernel-matrix-based SPD representation when built upon deep local descriptors and the superiority of the proposed DeepKSPD network. Future work will further explore the effectiveness of this network on other recognition tasks and develop the SPD representations in other forms.

\section{Appendix: Proof for Eq.(8) in main text}\label{sec:Appendix-A}
According to Eq.(2) and following the argument in Eq.(5), it can be shown that
{\small
\begin{equation}
\begin{aligned}
 \delta{J}=\left\langle\mathrm{vec}\left(\frac{\partial{J_2}}{\partial{\bm E}}\right),\mathrm{vec}({\delta{\bm E}})\right\rangle = \mathrm{trace}\left(\left(\frac{\partial{J_2}}{\partial{\bm E}}\right)^T{\delta{\bm E}}\right),\nonumber
\end{aligned}
\end{equation}
}%
where $\mathrm{vec}(\cdot)$ denotes the vectorization of a matrix and $\langle\cdot,\cdot\rangle$ denotes the inner product. Combining this result with $\delta{\bm E}=({\bm I}\circ\delta{\bm A}){\bm 1}+{\bm 1}^T({\bm I}\circ\delta{\bm A})^T-2\delta{\bm A}$ in Eq.(4), it can be obtained that
{\small
\begin{equation}
\begin{aligned}
& \mathrm{trace}\left(\left(\frac{\partial{J_2}}{\partial{\bm E}}\right)^T{\delta{\bm E}}\right) 
= \mathrm{trace}\left(\left(\frac{\partial{J_2}}{\partial{\bm E}}\right)^T{\left(({\bm I}\circ\delta{\bm A}){\bm 1}+{\bm 1}^T({\bm I}\circ\delta{\bm A})^T-2\delta{\bm A}\right)}\right).\nonumber
\end{aligned}
\end{equation}
}%
Keeping applying the identity that $\mathrm{trace}({\bm A}^{T}({\bm B}\circ{\bm C}))=\mathrm{trace}(({\bm B}\circ{\bm A})^T{\bm C})$, we can have
{\small
\begin{equation}
\begin{aligned}
& \mathrm{trace}\left(\left(\frac{\partial{J_2}}{\partial{\bm E}}\right)^T{\delta{\bm E}}\right) 
 = \mathrm{trace}\left(\left({\bm I}\circ\left(\left(\frac{\partial{J_2}}{{\partial{\bm E}}}+\left(\frac{\partial{J_ 2}}{{\partial{\bm E}}}\right)^T\right){\bm 1}^T\right)-2\frac{\partial{J_2}}{{\partial{\bm E}}}\right)^T\delta{\bm A}\right).\nonumber
\end{aligned}
\end{equation}
}%
Because we know $\delta{J}$ can also be expressed as $\mathrm{trace}\left(\left(\frac{\partial{J_1}}{\partial{\bm A}}\right)^T{\delta{\bm A}}\right)$ and the last result is valid for any $\delta{\bm A}$, it can be obtained that
{\small
\begin{equation}
\begin{aligned}
 \frac{\partial{J_{1}}}{{\partial{\bm A}}} = {\bm I}\circ\left(\left(\frac{\partial{J_{2}}}{{\partial{\bm E}}}+\left(\frac{\partial{J_{2}}}{{\partial{\bm E}}}\right)^T\right){\bm 1}^T\right)-2\frac{\partial{J_{2}}}{{\partial{\bm E}}}.\nonumber
\end{aligned}
\end{equation}
}%
This gives rise to the first half of Eq.(8).

Again, combining $\delta{J}=\mathrm{trace}\left(\left(\frac{\partial{J_1}}{\partial{\bm A}}\right)^T{\delta{\bm A}}\right)$ with $\delta{\bm A}=\left(\delta{\bm X}\right){\bm X}^T+{\bm X}\left(\delta{\bm X}\right)^T$ in Eq.(4), it can be obtained that
{\small
\begin{equation}
\begin{aligned}
& \mathrm{trace}\left(\left(\frac{\partial{J_1}}{\partial{\bm A}}\right)^T{\delta{\bm A}}\right) 
= \mathrm{trace}\left(\left(\frac{\partial{J_1}}{\partial{\bm A}}\right)^T\left(\left(\delta{\bm X}\right){\bm X}^T+{\bm X}\left(\delta{\bm X}\right)^T\right)\right).\nonumber
\end{aligned}
\end{equation}
}%
Applying the identities that $\mathrm{trace}({\bm {ABC}})=\mathrm{trace}({\bm {CAB}})$ and $\mathrm{trace}({\bm {ABC}})=\mathrm{trace}(({\bm {ABC}})^T)$, we can obtain
{\small
\begin{equation}
\begin{aligned}
& \mathrm{trace}\left(\left(\frac{\partial{J_1}}{\partial{\bm A}}\right)^T{\delta{\bm A}}\right) = \mathrm{trace}\left(\left(\left(\frac{\partial{J_1}}{\partial{\bm A}}+\left(\frac{\partial{J_1}}{\partial{\bm A}}\right)^T\right){\bm X}\right)^T\delta{\bm X}\right).\nonumber
\end{aligned}
\end{equation}
}%
Because we know $\delta{J}$ can also be expressed as $\mathrm{trace}\left(\left(\frac{\partial{J}}{\partial{\bm X}}\right)^T{\delta{\bm X}}\right)$ and the last result is valid for any $\delta{\bm X}$, it can therefore be obtained that
{\small
\begin{equation}
\begin{aligned}
 \frac{\partial{J}}{{\partial{\bm X}}} = \left(\frac{\partial{J_1}}{\partial{\bm A}}+\left(\frac{\partial{J_1}}{\partial{\bm A}}\right)^T\right){\bm X}.\nonumber
\end{aligned}
\end{equation}
}%
This gives rise to the second half of Eq.(8).

In addition, $\frac{\partial{J}}{{\partial{\theta}}}$ can be derived in a similar manner. As previous, $\delta{J}$ can be equally written as
{\small
\begin{equation}
\begin{aligned}
\delta{J}=\mathrm{trace}\left(\left(\frac{\partial{J_{3}}}{{\partial{\bm K}}}\right)^T{\delta{\bm K}}\right),\quad\mathrm{and}\quad\delta{J}=\frac{\partial{J}}{{\partial{\theta}}}\cdot{\delta{\theta}},\nonumber
\end{aligned}
\end{equation}
}%
where $\theta$ is the width of the Gaussian kernel, a scalar. It is not difficult to see that by regarding ${\bm E}$ as constant, ${\delta{\bm K}} = (-{\bm K}\circ{\bm E})\cdot{\delta\theta}$. Therefore, it can be obtained that
{\small
\begin{equation}
\begin{aligned}
\mathrm{trace}\left(\left(\frac{\partial{J_{3}}}{{\partial{\bm K}}}\right)^T{\delta{\bm K}}\right) &=& 
\mathrm{trace}\left(\left(\frac{\partial{J_{3}}}{{\partial{\bm K}}}\right)^T\left((-{\bm K}\circ{\bm E})\cdot{\delta\theta}\right)\right)\\~\nonumber
&=&\mathrm{trace}\left(\left(\frac{\partial{J_{3}}}{{\partial{\bm K}}}\right)^T(-{\bm K}\circ{\bm E})\right)\cdot{\delta\theta}.\nonumber
\end{aligned}
\end{equation}
}%
Combining with the last equation, we have
{\small
\begin{equation}
\begin{aligned}
 \frac{\partial{J}}{{\partial{\theta}}} = \mathrm{trace}\left(\left(\frac{\partial{J_3}}{{\partial{\bm K}}}\right)^T\left(-{\bm K}\circ{\bm E}\right)\right).\nonumber
\end{aligned}
\end{equation}
}% 

\section{Appendix: Visualization of feature maps learned by DeepKSPD network}\label{sec:Appendix-C} 
\begin{tabular}{|c|c|c|c|}
      \includegraphics[width=\mywidth]{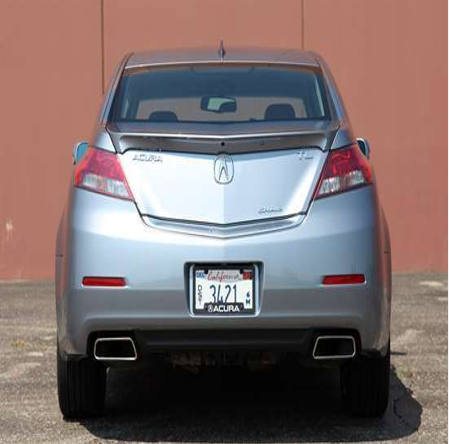} & \includegraphics[width=\mywidth]{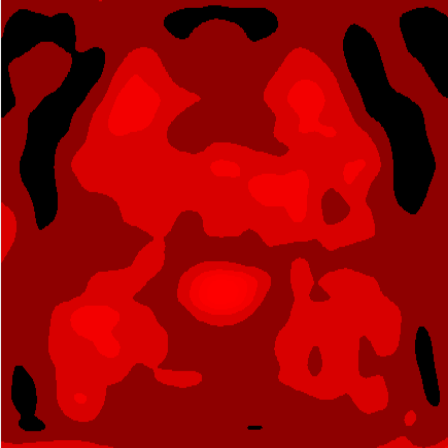} & \includegraphics[width=\mywidth]{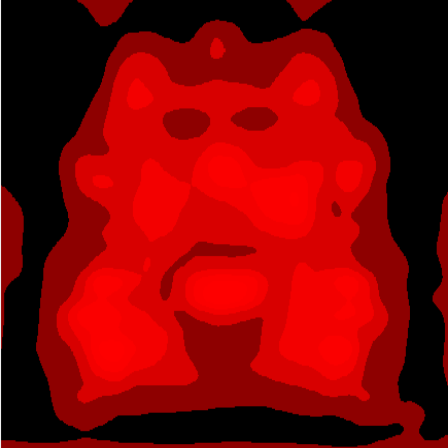} & \includegraphics[width=\mywidth]{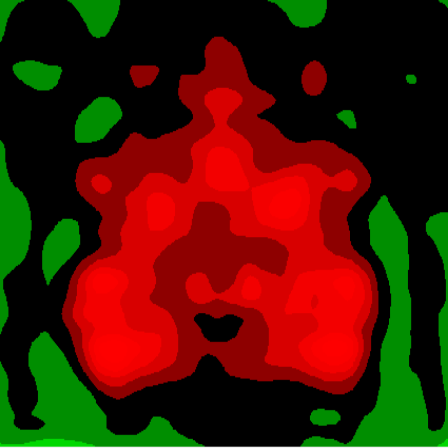} \\
      \includegraphics[width=\mywidth]{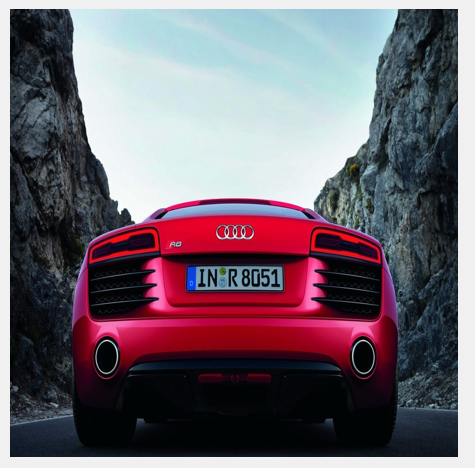} & \includegraphics[width=\mywidth]{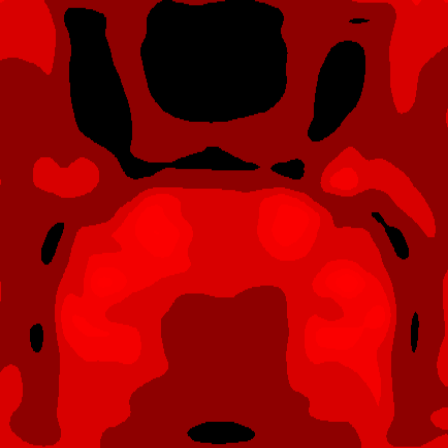} & \includegraphics[width=\mywidth]{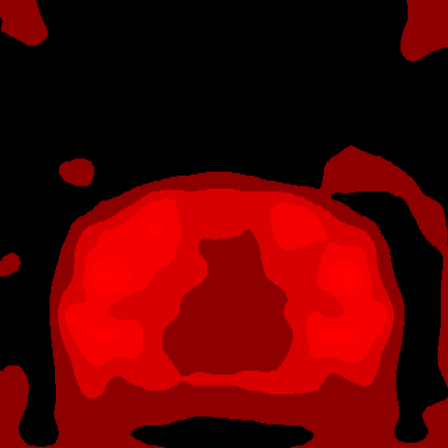} & \includegraphics[width=\mywidth]{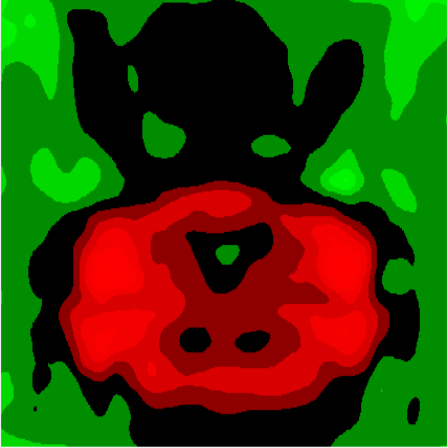} \\
            \includegraphics[width=\mywidth]{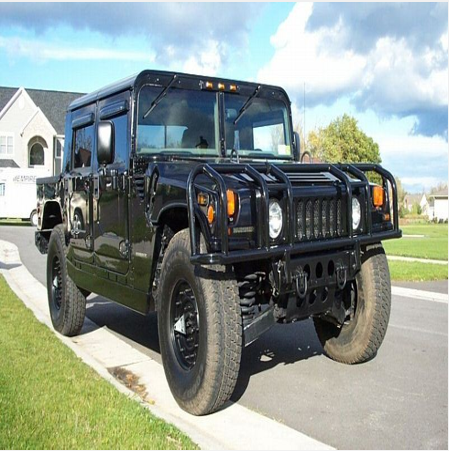} & \includegraphics[width=\mywidth]{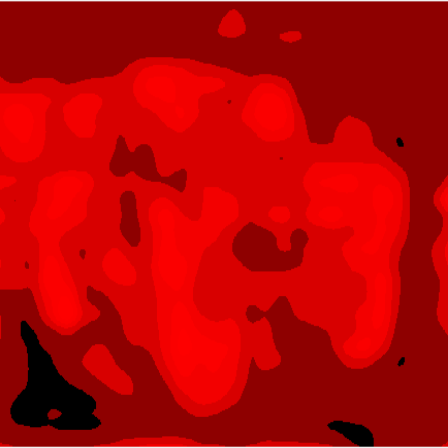} & \includegraphics[width=\mywidth]{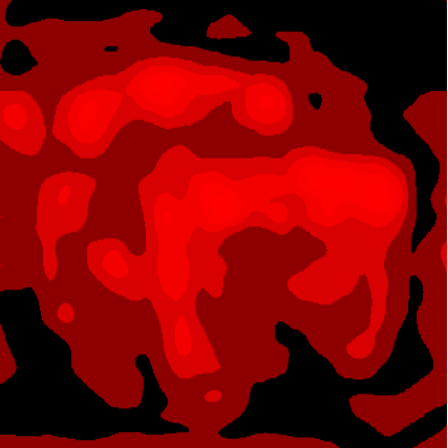} & \includegraphics[width=\mywidth]{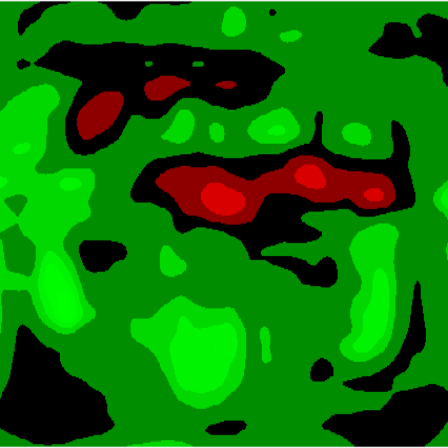} \\
      (a) Input image & (b) Before learning & (c) After learning & (d) Difference \\
\end{tabular} \vspace{2mm}

To gain more insight into the proposed DeepKSPD network, we visualize the activation feature maps (accumulated along the depth dimension) obtained with and without DeepKSPD learning. In the following figure, the four columns correspond to 1) the original input image; 2) the accumulated activation feature maps before learning (obtained from pretrained VGG-$19$ network); 3) the accumulated activation feature maps after learning (obtained from the trained DeepKSPD network); and 4) the difference between the two previous maps, where red color indicates increase and green color indicates decrease. 

As seen, the activations in the feature maps learned by DeepKSPD are generally enhanced on the body of the cars while reduced on the surroundings that are less relevant for car recognition. This shows that in the presence of the kernel-matrix-based SPD representation block, the DeepKSPD network is able to learn features that are meaningful from the perspective of recognition. This provides additional support to the excellent performance observed for DeepKSPD. 

\bibliographystyle{abbrv}
\bibliography{ms}

\end{document}